\documentclass[letterpaper, 10pt, conference]{ieeeconf}      

\IEEEoverridecommandlockouts 
\usepackage{cite}

\usepackage{amsmath,amssymb,amsfonts}
\usepackage{graphicx}
\usepackage{textcomp}

\DeclareMathOperator*{\argmin}{arg\,min}
\usepackage[table,xcdraw]{xcolor}
\usepackage{authblk}
\let\labelindent\relax
\usepackage{enumitem}
\usepackage{hyperref}
\usepackage{breqn}
\usepackage{algpseudocode}
\usepackage[linesnumbered,ruled,lined]{algorithm2e}
\usepackage{subfigure}

\usepackage{array}
\usepackage{makecell}
\usepackage{tabularray}

\usepackage{svg}

\allowdisplaybreaks

\begin{document}

\title{ Learning Sampling Distribution and Safety Filter for Autonomous Driving with VQ-VAE and Differentiable Optimization}
\author{Simon Idoko, Basant Sharma, Arun Kumar Singh\thanks{All authors are with the University of Tartu. This research was in part supported by financed by European Social Fund via ICT program measure, grants PSG753 from Estonian Research Council and collaboration project LLTAT21278 with Bolt Technologies. Our code is available at \url{https://github.com/cisimon7/VQOptMain}.
Emails: cisimon7@gmail.com, aks1812@gmail.com} 
}
\maketitle


\begin{abstract} 
Sampling trajectories from a distribution followed by ranking them based on a specified cost function is a common approach in autonomous driving. Typically, the sampling distribution is hand-crafted (e.g a Gaussian, or a grid). Recently, there have been efforts towards learning the sampling distribution through generative models such as Conditional Variational Autoencoder (CVAE). However, these approaches fail to capture the multi-modality of the driving behaviour due to the Gaussian latent prior of the CVAE. Thus, in this paper, we re-imagine the distribution learning through vector quantized variational autoencoder (VQ-VAE), whose discrete latent-space is well equipped to capture multi-modal sampling distribution. The VQ-VAE is trained with demonstration data of optimal trajectories. We further propose a differentiable optimization based safety filter to minimally correct the VQ-VAE sampled trajectories to ensure collision avoidance. We use backpropagation through the optimization layers in a self-supervised learning set-up to learn good initialization and optimal parameters of the safety filter. We perform extensive comparisons with state-of-the-art CVAE-based baseline in dense and aggressive traffic scenarios and show a reduction of up to $12$ times in collision-rate while being competitive in driving speeds.  
\end{abstract}

\begin{figure}[h!]
\centering
 \includegraphics[scale=0.08]{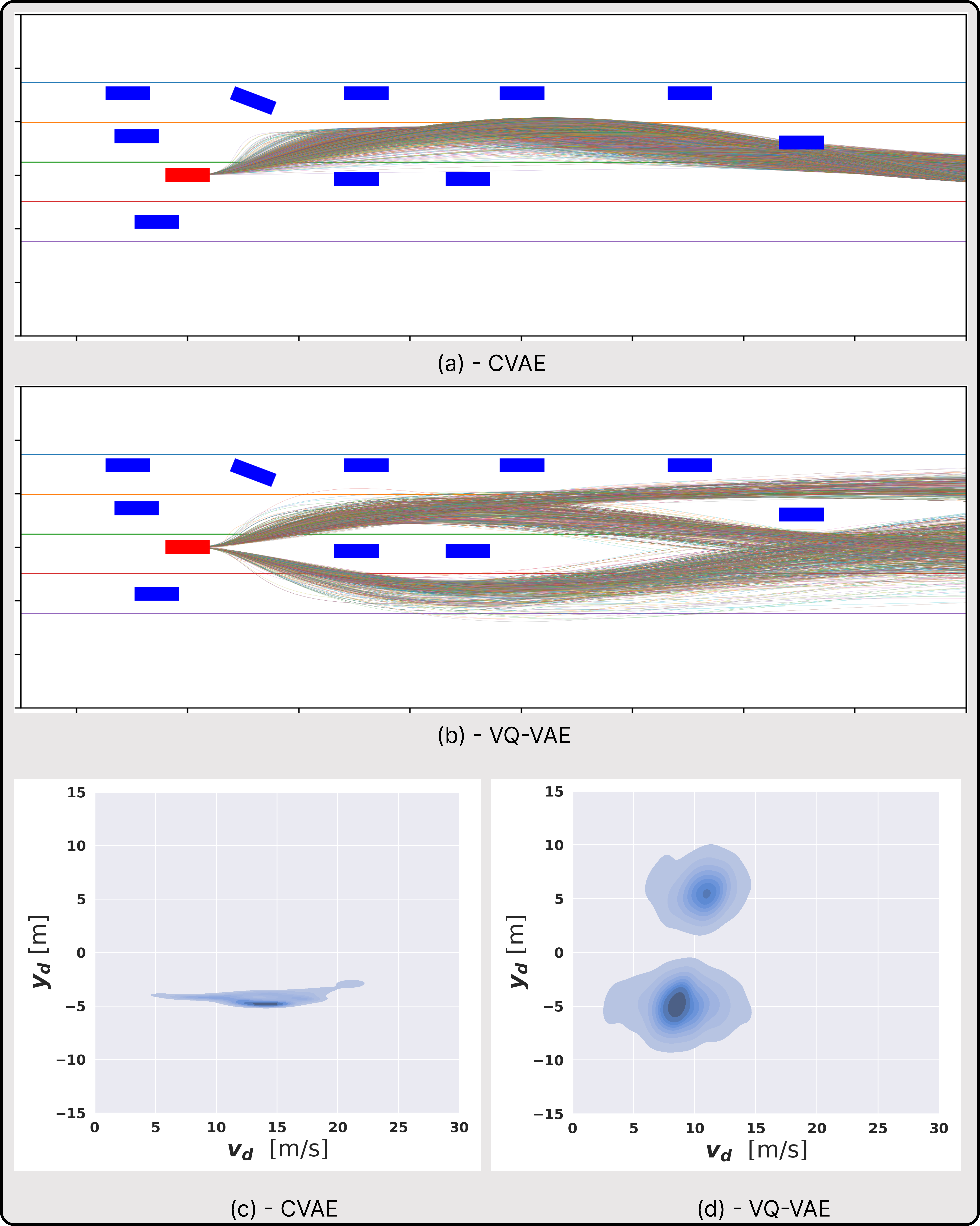}
\caption{Comparison between trajectory distribution sampled from CVAE (a) and VQ-VAE (b). As can be seen, VQ-VAE shows higher diversity and multi-modality in the sampled trajectories. This can be further validated by Fig. (c) and (d) which show the Kernel Density Estimation plots of the forward velocity ($v_d$) and lateral-offset ($y_d$) associated with each sampled trajectory.}
\label{vqvae_pipeline_teaser}
\vspace{-0.6cm}
\end{figure}

\section{Introduction}
Trajectory sampling is a conceptually simple approach that has found widespread adoption in autonomous driving community \cite{fernet_planner}, \cite{moller2024frenetix}. As the name suggests, the process involves sampling trajectories from a distribution and evaluating their utility based on a specified cost function. The least cost trajectory from the sample is chosen for execution. The sampling itself can be encoded in the form of interpretable parameters. For example, instead of trajectories, we can sample set-points for velocity and lateral offset for the vehicle and pass it through a simple quadratic program to obtain the resulting trajectory samples \cite{singh2023bi}. This encoding has the advantage of assigning some physical interpretation to each sampled trajectory.

Irrespective of the exact strategy, the underlying sampling distribution is often hand-crafted, e.g in the form of a Gaussian or a pre-fixed grid \cite{adajania2022multi}, \cite{moller2024frenetix}. Recently, there has been efforts towards learning the sampling distribution \cite{singh2023bi}, \cite{shrestha2023end}. The distribution itself is represented in the form of a Conditional Variational AutoEncoder (CVAE) \cite{kingmaauto} and can be conditioned on the environment observations. These cited approaches have shown superior performance compared to hand-crafted approaches. In this paper, we take this line of research further to solve one of the fundamental bottlenecks of CVAE, namely the posterior collapse. Intuitively, it refers to the inadequacy of CVAE in capturing multi-modal distributions, similar to what is commonly encountered in autonomous driving. This in turn, can be attributed to the Gaussian latent prior of the CVAE.

In this paper, we present a novel improvement based on Vector-Quantized Variational Autoencoder (VQ-VAE) \cite{oord2017neural}, whose discrete latent space is better suited to capturing multi-modal distribution. For example, the discrete latent space can capture the different homotopies of the optimal driving trajectories (see Fig.\ref{vqvae_pipeline_teaser}). Our key algorithmic contributions along with their benefits are summarized below

\noindent \textbf{Algorithmic Contribution:} We train a VQ-VAE using multi-modal demonstration of optimal trajectories to learn the underlying discrete latent space. We also train a PixelCNN \cite{van2016conditional} to sample from the learned latent space while conditioning it on the observation. We embed a differentiable QP within the VQ-VAE decoder to generate an intermediate interpretable representation of the each sampled trajectory from the VQ-VAE in terms of velocity and lateral-offset setpoints. 

We show that while VQ-VAE based trajectory sampling is enough for collision-free navigation in low density traffic, more complicated scenarios require explicit consideration of collision avoidance and lane boundary constraints. With this motivation, we propose an optimization-based safety filter modeled in terms of barrier function \cite{li2021safe}. The parameters of the filter along with good initialization for the underlying optimizer is learned in a self-supervised manner. We show how reformulations of barrier constraints can be exploited to simplify the differentiation through the safety filter optimization layer.  

\noindent \textbf{State-of-the-Art Performance:} We compare our approach with recent CVAE-based baseline presented in \cite{shrestha2023end} which has shown impressive performance in dense traffic scenarios. We show that our VQ-VAE pipeline achieves up to 12 times reduction in collision-rate over \cite{shrestha2023end}, while being competitive in achieved forward velocity. We further show that the discrete latent prior of VQ-VAE ensures superior diversity in sampled trajectories. As a result, a near-perfect performance is achieved in low traffic scenarios even without the computationally demanding safety layer. Finally, we show that our approach also maintains good performance even at reduced computation and sampling budget.

\section{Mathematical Preliminaries}
\subsubsection*{Symbols and Notation} 
Scalars will be denoted by normal font lowercase letters, vectors by bold font lowercase letters, and matrices by uppercase bold font letters. The superscript $T$ will indicate the transpose operation applied to either a matrix or a vector.



\begin{figure*}[h!]
\centering
 \includegraphics[scale=0.5]{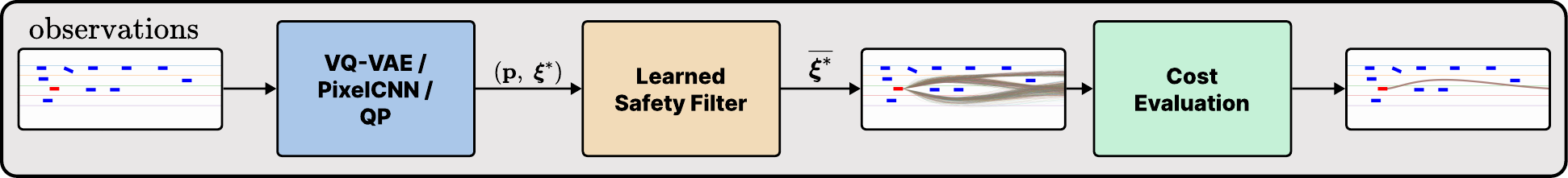}
\caption{Our overall pipeline that consists of sampling from a learnd VQ-VAE and passing the samples to a learned safety filter. This is followed by cost evaluation on the trajectory samples and the selection of the best trajectory.}
\label{pipeline}
\vspace{-0.5cm}
\end{figure*}
\subsection{Frenet Frame and Trajectory Parametrization}

\noindent We assume access to a lane center-line which allows us to perform motion planning in the so-called Frenet frame. In this set-up, the $X$ and $Y$ axes of the Frenet-frame are aligned with the longitudinal and lateral motion of the ego-vehicle. We parametrize the positional space ($x[k], y[k]$) of the ego-vehicle  in the Frenet frame at any time instant $k$ in terms of polynomials:

\vspace{-0.3cm}

\small
\begin{align}
    \begin{bmatrix}
        x[0],x[1], \dots, x[k] 
    \end{bmatrix} = \textbf{W}\textbf{c}_{x},
     \begin{bmatrix}
        y[0], y[1], \dots, y[k] 
    \end{bmatrix} = \textbf{W}\textbf{c}_{y},
    \label{param}
\end{align}
\normalsize

\noindent where, $\textbf{W}$ is a matrix formed with time-dependent polynomial basis functions and ($\textbf{c}_{x}, \textbf{c}_{y}$) are the coefficients of the polynomial. We can also express the derivatives in terms of $\dot{\textbf{W}}, \ddot{\textbf{W}}$.


\subsection{Trajectory Sampling Via Setpoints}

\noindent Instead of trajectories, we follow the intuition of \cite{hoel_rl_behavior}, \cite{shrestha2023end} and sample set-points for forward velocity and lateral-offset from the center-line. These are converted to a trajectory distribution by solving the following optimization problem. 

\small
\begin{subequations}
\begin{align}
  \min  \sum_k c_{s} +c_{l}+c_v\label{cost} \\
    (x^{(o)}[0],  y^{(o)}[0]) = \textbf{b}_0, (x^{(o)}[n],  y^{(o)}[n]) = \textbf{b}_f \label{boundary_cond}
\end{align}
\end{subequations}
\normalsize
\vspace{-0.5cm}
\small
\begin{subequations}
\begin{align}
    c_{s} (\ddot{x}[k], \ddot{y}[k]) = \ddot{x}[k]^2+\ddot{y}[k]^2\\
    c_{l}(\ddot{y}[k], \dot{y}[k]) = (\ddot{y}[k]-\kappa_p(y[k]-y_d)-\kappa_v\dot{y}[k])^2\\
    c_v(\dot{x}[k], \ddot{x}[k]) = (\ddot{x}[k]-\kappa_p(\dot{x}[k]-v_d))^2
\end{align}
\end{subequations}
\normalsize

\noindent The first term $c_s(.)$ in the cost function \eqref{cost} ensures smoothness in the planned trajectory by penalizing high accelerations at discrete time instants. The last two terms ($c_l(.), c_v(.)$) model the tracking of lateral offset ($y_{d}$) and forward velocity $(v_{d})$ set-points respectively and is inspired from works like \cite{hoel_rl_behavior}. For the former, we define a Proportional Derivative (PD) like tracking with gain $(\kappa_p, \kappa_v)$. It induces lateral accelerations that will make the ego-vehicle converge to the $y_d$. The derivative terms in $c_l$ minimize oscillations while converging to the desired lateral offset. For velocity tracking, we only use a proportional term.  Equality constraints \eqref{boundary_cond} ensures boundary conditions on  the $o^{th}$ derivative of the planned trajectory. We use $o= \{0, 1, 2\}$ in our formulation.

Optimization \eqref{cost}-\eqref{boundary_cond} can be converted into the following QP using the trajectory parameterization of \eqref{param}.

\small
\begin{subequations}
\begin{align}
    \boldsymbol{\xi}^* = \argmin_{\boldsymbol{\xi}} \frac{1}{2}\boldsymbol{\xi}^T\textbf{Q}\boldsymbol{\xi}+\textbf{q}^T(\textbf{p})\boldsymbol{\xi}, \label{lower_cost_reform}  \\
    \textbf{A}\boldsymbol{\xi} = \textbf{b}\label{lower_eq_reform} 
\end{align}
\end{subequations}
\normalsize
\noindent where $\boldsymbol{\xi} = (\textbf{c}_x, \textbf{c}_y)$ and $\textbf{p} = (v_d, y_d)$.

\section{Main Results}

\noindent The complete pipeline of our model is illustrated in fig \ref{pipeline}. Our goal is to generate a distribution of safe trajectories corresponding to a given observation $\mathcal{O}$. This is accomplished through the different blocks as show in fig \ref{pipeline}. Firstly, we sample from a learned VQ-VAE to obtain a distribution of velocity and lateral-offset set points ($\textbf{p}$), and the associated trajectories. The sampled trajectories are then processed by a learned safety filter block, which ensures that they conform to collision and lane boundary constraints. Following this safety filtration, the trajectories are evaluated by a cost function to identify the optimal (least-cost) trajectory. The VQ-VAE and the safety filter are trained separately in a heirarchical manner. We describe the main building blocks next.

\subsection{Learning with VQ-VAE and Differentiable QP block}
\begin{figure*}[h!]
\centering
 \includegraphics[scale=0.5]{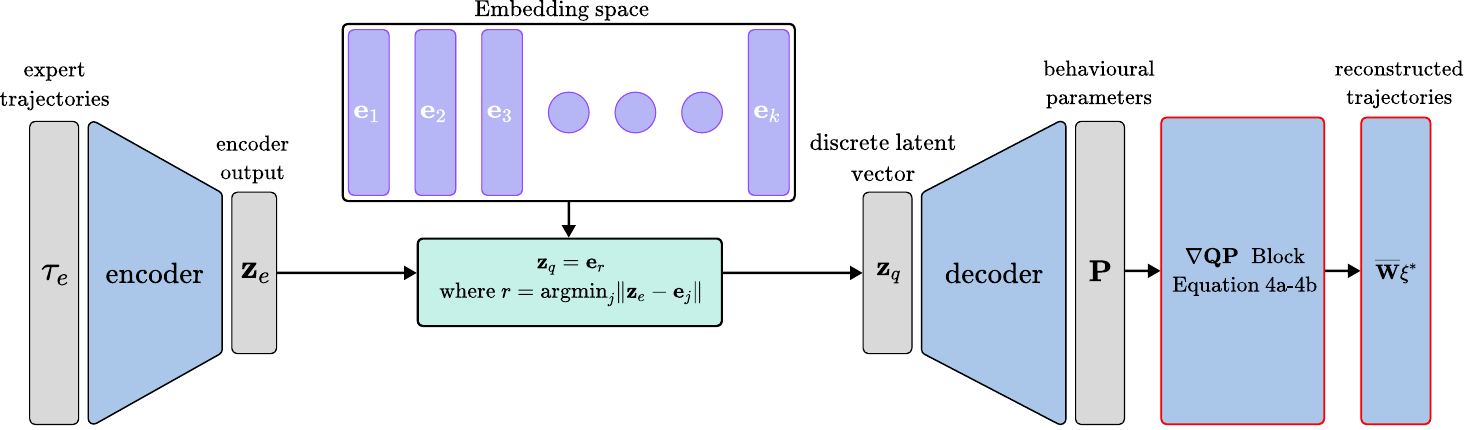}
\caption{VQVAE with a differentiable QP block for learning discrete latent prior over the forward velocity $v_d$, lateral offset $y_d$ and the associated trajectory.  }
\label{vqvae_pipeline}
\vspace{-0.2cm}
\end{figure*}

\noindent Our VQ-VAE pipeline is shown in Fig.\ref{vqvae_pipeline} and adapts \cite{oord2017neural} for trajectory inputs. The expert trajectories are mapped by an encoder to a continuous latent space $\mathbf{Z}_e \in \mathbb{R}^{L \times D}$, where $L$ is the number of latent vectors and $D$ is their individual dimensionality. For ease of notation, we represent the $i^{th}$  latent vector (row) of  $\mathbf{Z}_e$ as $\mathbf{z}_{e, i}$. The core idea of VQ-VAE lies in the discretization of the latent space. i.e mapping continuous $\mathbf{Z}_e$ to discrete $\textbf{Z}_q$. To this end, we introduce a latent embedding space $\mathbf{E} \in \mathbb{R}^{K \times D}$ consisting of $K$ discrete vectors $\textbf{e}_j$ with dimension $D$. Subsequently, we define the $i^{th}$ latent vector (row) of $\textbf{Z}_q$ as the $j^{th}$ $\textbf{e}_j$ vector closest to $\mathbf{z}_{e, i}$. This assignment is performed by the nearest neighbour optimization problem defined in \eqref{nearest}

\small
\begin{multline}
    \boldsymbol{z}_{q,i} = \boldsymbol{e}_r, \text{ where } r = \arg \min_j || \boldsymbol{z}_{e, i} - \boldsymbol{e}_j ||_2^2   \\
    \text{for } i \in \{1, 2, ..., L\} 
    \label{nearest}
\end{multline} 
\normalsize

The decocder of VQ-VAE takes in $\textbf{Z}_q$ and produces set-points $\mathbf{p}$, which is then passed through a QP layer defined by \eqref{lower_cost_reform}-\eqref{lower_eq_reform} to obtain the reconstruction of the expert trajectory $\boldsymbol{\xi}^*$. The VQ-VAE is trained using the loss function defined in \eqref{vq_loss} which has three components.  The first part is the reconstruction loss which makes the entire pipeline learn to reconstruct the input expert trajectory. The typical backpropagation cannot trace the gradient through the  non-differentiable nearest neighbour assignment \eqref{nearest}, which in turn poses problem in jointly training the encoder, decoder and the embedding space vectors. To counter this, straight-through gradient estimation method proposed in 
\cite{oord2017neural} is used, which in turn necessitates adding the last two terms in the loss function \eqref{vq_loss}. The second term in the loss function employs the $l_2$ error to guide the movement of embedding vectors $\boldsymbol{e}_i$ towards the encoder outputs $\boldsymbol{z}_e$, thereby updating the embedding space vectors parameters. And the last term seeks to make the encoder commit to the embedding space vectors. In other words, it prevents the encoder output to keep getting assigned to a different nearest neighbour in each forward pass. 

\small
\begin{multline}
    \mathcal{L}_{VQVAE} = \| \overline{\boldsymbol{W}} \boldsymbol{\xi}^* - \boldsymbol{\tau}_e \|_2^2 \;+\; \| \text{sg}[\mathbf{Z}_e] - \mathbf{E}\|_2^2 \\
    \;+\;  \beta \| \mathbf{Z}_e(x) - \text{sg}[\mathbf{E}]\|_2^2
    \label{vq_loss}
\end{multline} 
\normalsize

\subsection{Conditional sampling of the discrete latent space using PixelCNN}
\noindent Let $\boldsymbol{h}_q$ denote a vector wherein each element corresponds to the codebook index of the vector $\boldsymbol{z}_{q,i}$ in $\boldsymbol{Z}_q$, with $i \in 1,2,...,L$. We want to generate different samples of $\boldsymbol{h}_q$ such that they can be mapped to the learned latent space of VQ-VAE, $\mathbf{Z}_q$. These in turn, can be mapped to velocity and lateral offset setpoints $(\mathbf{p})$ and the associated trajectories through the learned decoder. Moreover, the $\boldsymbol{h}_q$ generation should be conditioned on the observations $\mathcal{O}$. To this end, we adapt the PixelCNN architecture proposed in \cite{van2016conditional}.

\begin{figure}[h!]
\centering
 \includegraphics[scale=0.5]{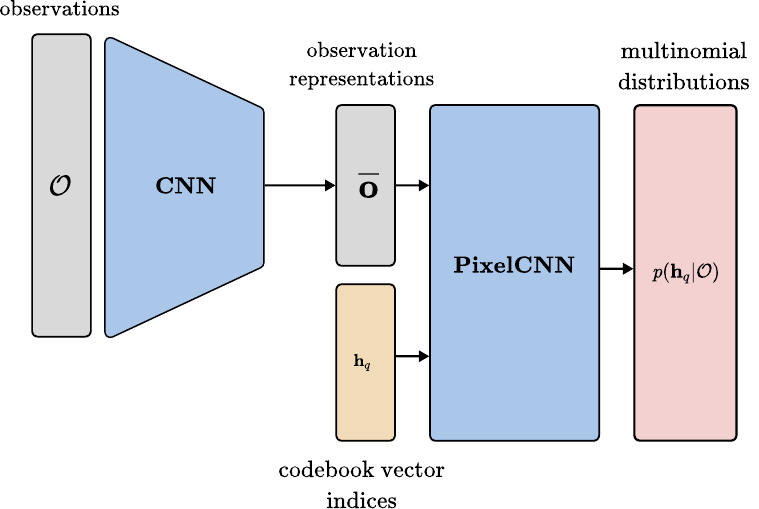}
\caption{The auto-regressive conditional PixelCNN architecture that is used to learn how to sample from the discrete latent space of a trained VQ-VAE.}
\label{pix_pipeline}
\vspace{-0.2cm}
\end{figure}

Our pipeline is shown in fig \ref{pix_pipeline}. We pass observation vector $\mathcal{O}$ through CNN layers to get feature vector $\overline{\mathbf{o}}$. During training, this representation, together with the ground truth $\mathbf{h}_q$ obtain from the trained VQ-VAE model is passed to the PixelCNN. The model then predicts the conditional distribution of the codebook indices in an autoregressive manner using a set of masked CNN layers as described in \cite{van2016conditional}. We train the generated distribution $ p(\boldsymbol{h}_q|\mathcal{O})$ using cross-entropy loss with the ground truth $\mathbf{h}_q$.


During the inferencing phase, we iniatialize $\mathbf{h}_q$ to zeros and recursively generate the element of $\mathbf{h}_q$ conditioned on the observation using the trained PixelCNN. This generates a multinomial distribution for each generated $\mathbf{h}_q$ element. We sample from this distribution at each step to construct the corresponding $\mathbf{h}_q$ vector which then passes through the embedding layer to get $\mathbf{Z}_q$.

\subsection{Learnable Safety Filter}\label{LSF}
\noindent Our safety filter is defined through the following optimization problem.
\small
\begin{align}
    \overline{\boldsymbol{{\xi}}}^{*}_j = \arg\min_{\overline{\boldsymbol{\xi}}^*_j} \frac{1}{2}\Vert \overline{\boldsymbol{\xi}}^*_j-\boldsymbol{\xi}^*_j\Vert_2^2 \label{projection_cost}\\
    \textbf{A}\overline{\boldsymbol{\xi}}^*_j = \textbf{b}, \qquad \textbf{g}(\overline{\boldsymbol{\xi}}^*_j, \boldsymbol{\gamma}) \leq  \textbf{0} \label{projection_const}
\end{align}
\normalsize

\noindent As can be seen, the safety filter takes in the $j^{th}$ trajectory generated by the VQ-VAE and projects them onto the constraint set. The equality constraints ensure that the trajectories satisfy the initial and terminal states. The inequality constraints model collision avoidance, lane-boundary constraints along with velocity and acceleration bounds. We defined the algebraic form of these entities in Appendix. But we highlight that the collision avoidance and lane-boundary constraints are designed using the barrier function approach of \cite{li2021safe}. In this context, $\boldsymbol{\gamma}$ represents the parameter of the barrier function.

The safety filter has one  explicit learnable parameter namely $\boldsymbol{\gamma}$. Additionally, we also aim to learn good initialization for optimization problem underlying the safety filter. 

\begin{figure}[h!]
\centering
 \includegraphics[scale=0.6]{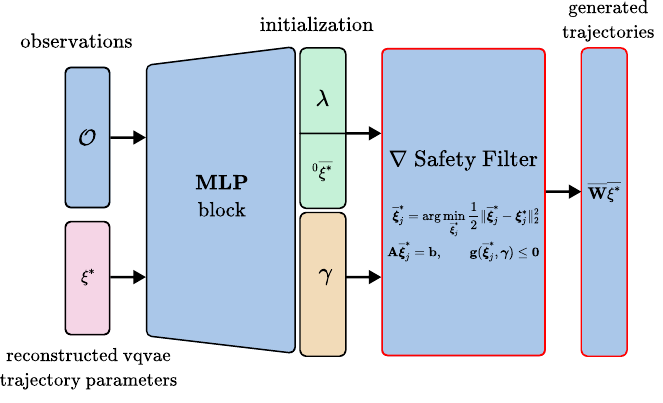}
\caption{Learned Safety filter that consists of an MLP augmented with an optimization layer. We train the safety filter in self-supervised setting to learn the parameters of inequality constraints along with good intialization for the optimization problem.}
\label{mlp_projection}
\vspace{-0.2cm}
\end{figure}

\noindent \textbf{Training Pipeline:} The overall architecture of safety filter is shown in Fig.\ref{mlp_projection}. A multi-layer perceptron (MLP) block takes in the observations $\mathcal{O}$ along with the trajectory generated by the VQ-VAE and outputs $\boldsymbol{\gamma}$ along with initialization $\boldsymbol{\lambda}$ (defined in Appendix \ref{appendix}), ${^0}\overline{\boldsymbol{\xi}}^*$ for the optimization solver. These predictions are then passed onto the safety filter. The MLP is trained in a self-supervised manner using the loss function defined in \eqref{mlp_loss}. The first term in the loss function minimizes the projection residual and is designed to enforce minimum possible correction to the input trajectory. The second term $c(\overline{\boldsymbol{\xi}^*})$ minimizes the constraint residuals of the safety filter. The constant $s$ is used to trade-off between the two loss terms.

\vspace{-0.3cm}
\small
\begin{align}
    \mathcal{L}_{mlp} = \| \boldsymbol{\xi}^* - \overline{\boldsymbol{\xi}^*} \|_2^2 \;+\; s \sum_{k} {c(\overline{\boldsymbol{\xi}^*})}
    \label{mlp_loss}
\end{align}
\normalsize

\noindent \textbf{Differentiation Through the Optimization Layer: } Training the safety filter requires differentiating through the optimization problem underlying the safety filter. There are two possible approaches for it namely implicit differentiation and algorithm unrolling \cite{pinedatheseus}. The former cannot be used in our case, as they are not suitable for learning initialization. More precisely, the initialization do not appear explicitly in the optimality conditions and thus, these cannot be learned using implicit differentiation. On the other hand, algorithm unrolling does not have this limitation but requires that every step of the optimization solver is differentiable. To this end, we propose a novel solution process that ensures differentiability across each numerical step. Detailed analysis can be found in Appendix \ref{appendix}, \eqref{am_alpha}-\eqref{am_xi}. Moreover, our solver is easily parallelizable over GPUs, does not require any matrix factorization and even has closed-form solution for many of the intermediate steps.



\newtheorem{remark}{Remark}\label{rem_1}

\section{Connections to Existing Works} \label{connections}
\subsubsection*{Learning Trajectory Sampling Distribution} A large number of recent works have attempted to learn good sampling distribution. The application domain covers both general motion planning as well as those specific for autonomous driving. For example, \cite{ichter2018learning} uses CVAE to approximate a good sampling distribution for motion planners. Along similar lines, \cite{sacks2023learning} learned sampling distribution to accelerate MPPI \cite{williams2017model} algorithm using normalizing flows \cite{papamakarios2021normalizing}. Our prior work used CVAE along with differentiable optimization layers \cite{singh2023bi}, \cite{shrestha2023end}. The fundamental difference between these cited works stems from our usage of discrete latent priors, which we believe better captures the multi-modality for autonomous driving trajectories. 

\subsubsection*{Differentiable Optimization Layers} The ability to differentiate optimization solvers have proved extremely useful for end-to-end learning \cite{amos2017optnet} for control. However, the greatest success of these class of approaches have come while embedding convex optimization problems within neural network pipeline. Algorithms for differentiating non-convex solvers is limited, for example, unconstrained non-linear least squares problem \cite{pinedatheseus}. Thus, current work develops its own custom differentiable optimization layers following our prior efforts \cite{shrestha2023end}. In particular, we reformulate the underlying constraints to achieve an optimization routine where each step is differentiable. Subsequently, we use algorithm unrolling to learn both good initialization as well as the explicit learnable parameters of our safety filter. 

\subsubsection*{Safety Filter: } Safety filter based on barrier function are extensive used in the existing works \cite{brunke2022safe}. Potentially, the parameters of the CBF can also be learned, for example through imitation learning \cite{xiao2023barriernet}. However, most existing safety filter, especially the ones that are learned are based on one-step planning formulated as a QP \cite{xiao2023barriernet}. In contrast, our proposed safety filter is based on trajectory optimization over a long horizon.



\section{Validation and benchmarking}
\noindent This section presents extensive simulation results to answer the following research questions.

\begin{itemize}
    \item \textbf{Q1} How does VQ-VAE compare with CVAE in capturing multi-modal trajectory distribution and how these respective approaches perform in complex driving scenarios.
    \item \textbf{Q2} How does the performance of VQ-VAE and CVAE scale with restriction in computational and sampling budget.
\end{itemize}

\begin{figure}[h!]
\centering
 \includegraphics[scale=0.2]{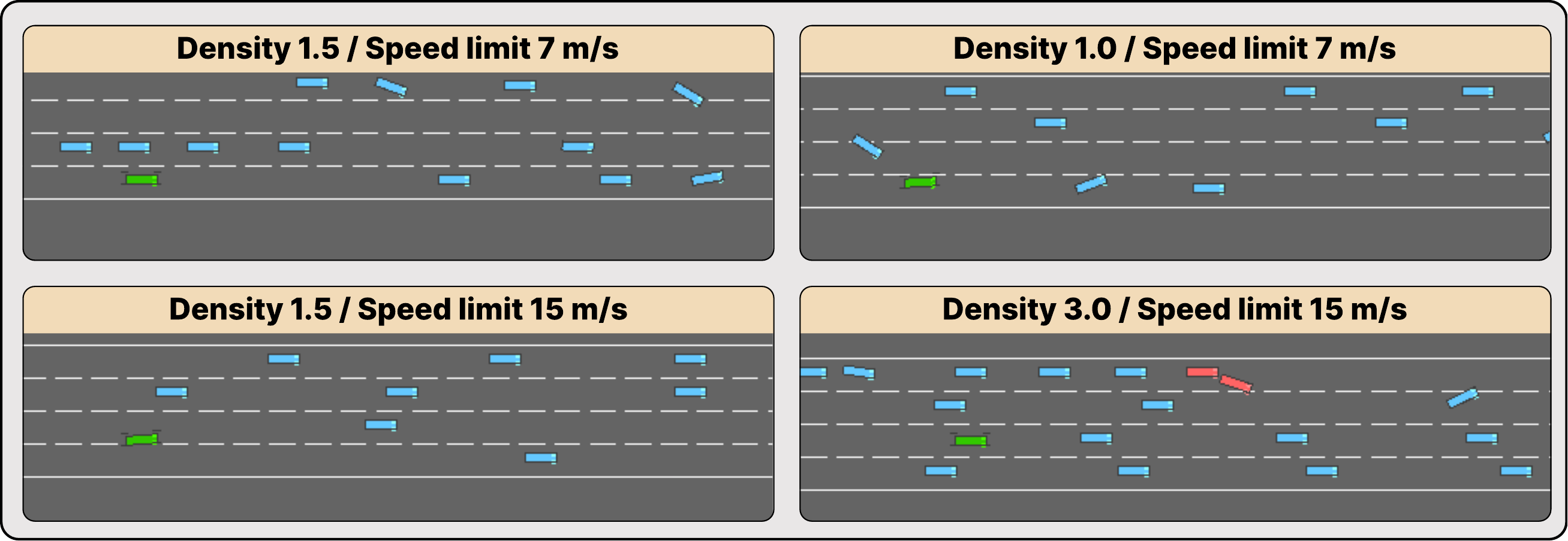}
\caption{Various traffic scenarios made from different densities and speed limit of neighboring vehicles used for benchmarking our approach with the CVAE-based baseline \cite{shrestha2023end}}
\label{scenarios}
\vspace{-0.2cm}
\end{figure}

\subsection{Implementation Details}
\noindent We developed our entire inferencing pipeline in Python with JAX \cite{jax} as the GPU-accelerated linear algebra back-end. The matrix $\textbf{W}$ in \eqref{param} is derived from a $10^{th}$ order polynomial. The VQ-VAE, PixelCNN and safety filter were trained in PyTorch. Our simulation framework is based on the Highway Environment (highway-env) simulator \cite{Leurent_An_Environment_for_2018}, where neighboring vehicles adhere to simple rule-based strategies for lateral and longitudinal control.


\subsubsection{Training of VQ-VAE and PixelCNN} The training details of the VQ-VAE and PixelCNN models have been elaborated in earlier sections. Each expert trajectory consists of 100 $x-y$ coordinate points. The observation vector $\mathcal{O}$, sized 55, consisted of (i) distance from the left and right lane boundaries, (ii) lateral and longitudinal velocities, (iii) ego-vehicle heading, and (iv) state information (x-y position, latitudinal and longitudinal velocities, and heading) of the ten closest obstacles. All position measurements are relative to the ego-vehicle position. During inference, the PixelCNN uses $\mathcal{O}$ from the simulator to yield vector samples of codebook indices $\boldsymbol{h}_q$ in the embedding space. These indices are then fed through the embedding layer to retrieve the discrete latent representation $\boldsymbol{Z}_q$. Subsequently, the VQ-VAE decoder first generates the corresponding velocity and lateral-offset vector $\textbf{p}$ and then the associated trajectories.
To gather multi-modal optimal trajectory demonstrations for training both the VQ-VAE and PixelCNN, we use the approach presented in \cite{rastgar2023gpu}

\subsubsection{Baseline}
We compare our VQ-VAE based approach with the CVAE-based baseline presented in \cite{shrestha2023end}, which has earlier demonstrated good performance in dense traffic scenarios. Both the CVAE and VQ-VAE models utilize identical observation vectors denoted as \( \mathcal{O} \) as input. Both the approaches also employ a safety filter. But the ones used in  \cite{shrestha2023end} does not have any explicit learnable parameter, while that proposed in current work learns the barrier function parameters through self-supervised learning. 

\subsubsection{Environments, Tasks, and Metrics}

The highway driving scenarios are depicted in Figure \ref{scenarios}. Utilizing the highway-env simulator, we are able to configure parameters such as vehicle density and average speed limit within each traffic scenario. Benchmarking was conducted across multiple scenarios, each evaluated using two distinct seeds, with the resulting metric averages being derived from these seeds.

The primary objective is collision-free navigation of the ego vehicle amidst traffic, while attempting to move as fast as possible. Consequently, the evaluation metric comprises two key components: (i) collision rate and (ii) average velocity attained within each episode, with collision rate taking precedence as the more critical measure.


\subsection{Qualitative Comparison Between VQ-VAE and CVAE}
\noindent Fig.\ref{vqvae_pipeline_teaser} (a) shows a scenario with static obstacles. The ego-vehicle is positioned in a way that multiple feasible trajectories are possible. However, CVAE-based approach of \cite{shrestha2023end} is only able to generate trajectory samples in one homotopy class. In sharp contrast, our VQ-VAE (Fig.\ref{vqvae_pipeline_teaser}(b)) can leverage its learned discrete latent space to generate more diverse trajectories. The core difference between the two models can be further understood by Fig.\ref{vqvae_pipeline_teaser}(c)-(d), which shows the Kernel Density Estimation (KDE) over the velocity and lateral-offset setpoints associated with the sampeled trajectories. While VQ-VAE generated samples exhibit sharp mutli-modality, a distinct posterior collapse can be viewed for the CVAE.


\subsection{Benchmarking Driving Performance with VQ-VAE and CVAE}
\noindent All tasks referred to in this section were performed across 50 different episodes and two different seeds. The collision rate measures the percentage number of crashed episodes in the 50 episode runs. For forward velocity, we present both mean and standard-deviation to capture the variation within episodes. 

\subsubsection{Performance Across Different Densities}
\begin{figure}[h!]
\centering
 \includegraphics[scale=0.28]{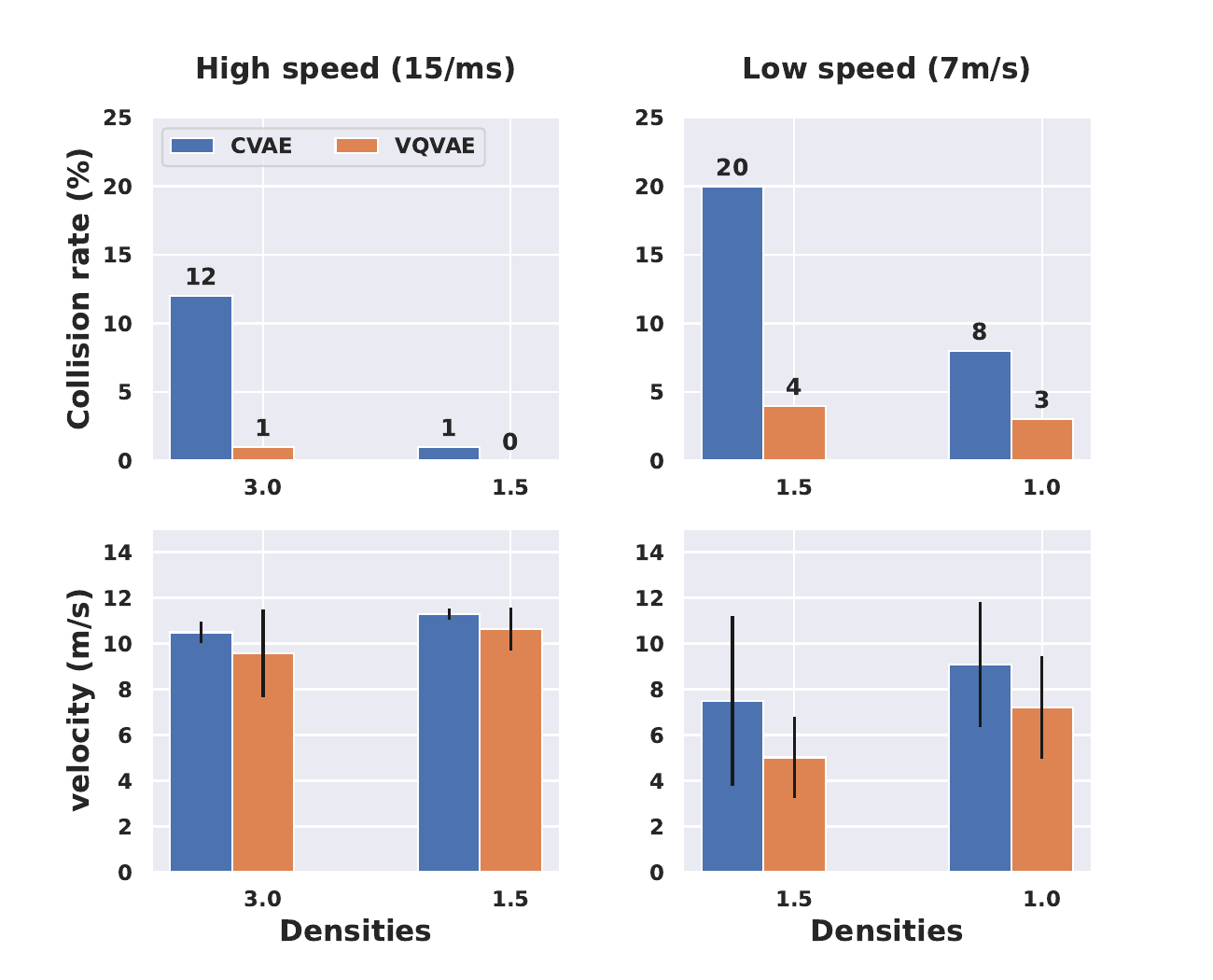}
\caption{Comparison of CVAE and VQ-VAE based models across different densities and speed limit of the neighboring vehicles. In both scenarios, the neighboring vehicles randomly chooses their forward velocity between zero and the speed limit. VQ-VAE massively outperforms the CVAE baseline in terms of collision-rate.}
\label{maxiter_speed}
\vspace{-0.2cm}
\end{figure}

Fig.\ref{maxiter_speed} compares the performance of CVAE based trajectory sampling of \cite{shrestha2023end} with our VQ-VAE pipeline. The former works well for low-traffic scenarios and where the speed limit of the neighboring vehicle is high creating large vacant spaces. As the density increases or the neighboring vehicles start moving slowly, \cite{shrestha2023end} demonstrates higher collision-rate. In contrast, the superior exploration properties of our VQ-VAE sampling allows for upto 12 times reduction (density 3 scenarios) in collision-rate over the CVAE baseline. Both the approaches result in similar velocity profiles.


\subsubsection{Performance Scaling with Computational Budget}
\begin{figure}[h!]
\centering
 \includegraphics[scale=0.27]{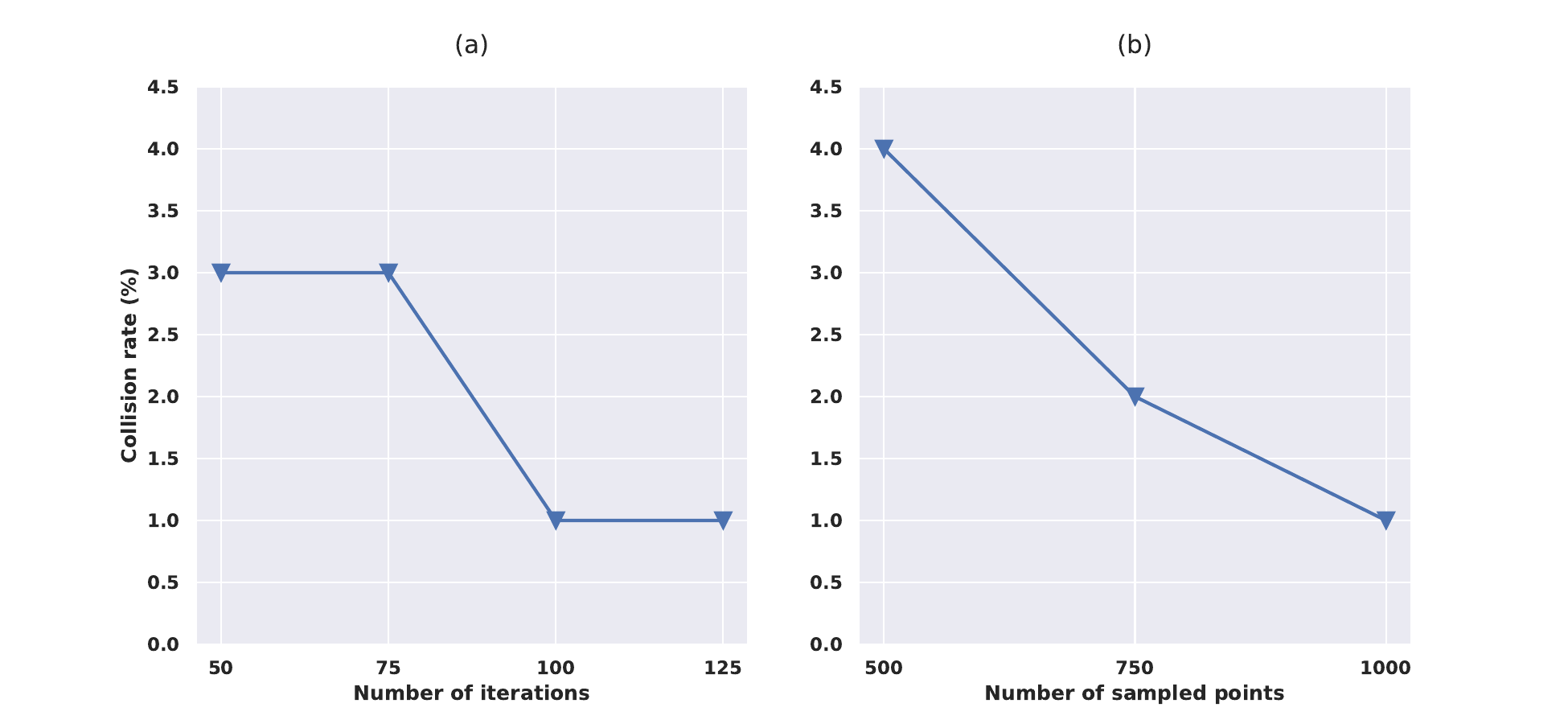}
\caption{(a) shows the comparison across different number of iterations of the safety filter. (b) shows the comparison across different number of sampled set points/trajectories.}
\label{maxiter_batch}
\vspace{-0.2cm}
\end{figure}

In this study, we examine the performance of the VQ-VAE model under reduced computational resources. Specifically, we focus on variations in the number of iterations 
of the safety layer optimization and the number of trajectories sampled from VQ-VAE. Fig. \ref{maxiter_batch} summarizes the results. We observe that the collision-rate is mere $3\%$ even when the number of iterations is reduced to 50. Similarly, while reducing the sample-size from 1000 to 750, the collision-rate has only increased to $4\%$.
It is worth pointing out that these experiments were conducted under conditions where the traffic density was set at 3.0 and the speed limit of neighboring vehicles was 15 m/s.

\subsection{Effect of Learned Safety Filter}

\begin{figure}[h!]
\centering
 \includegraphics[scale=0.28]{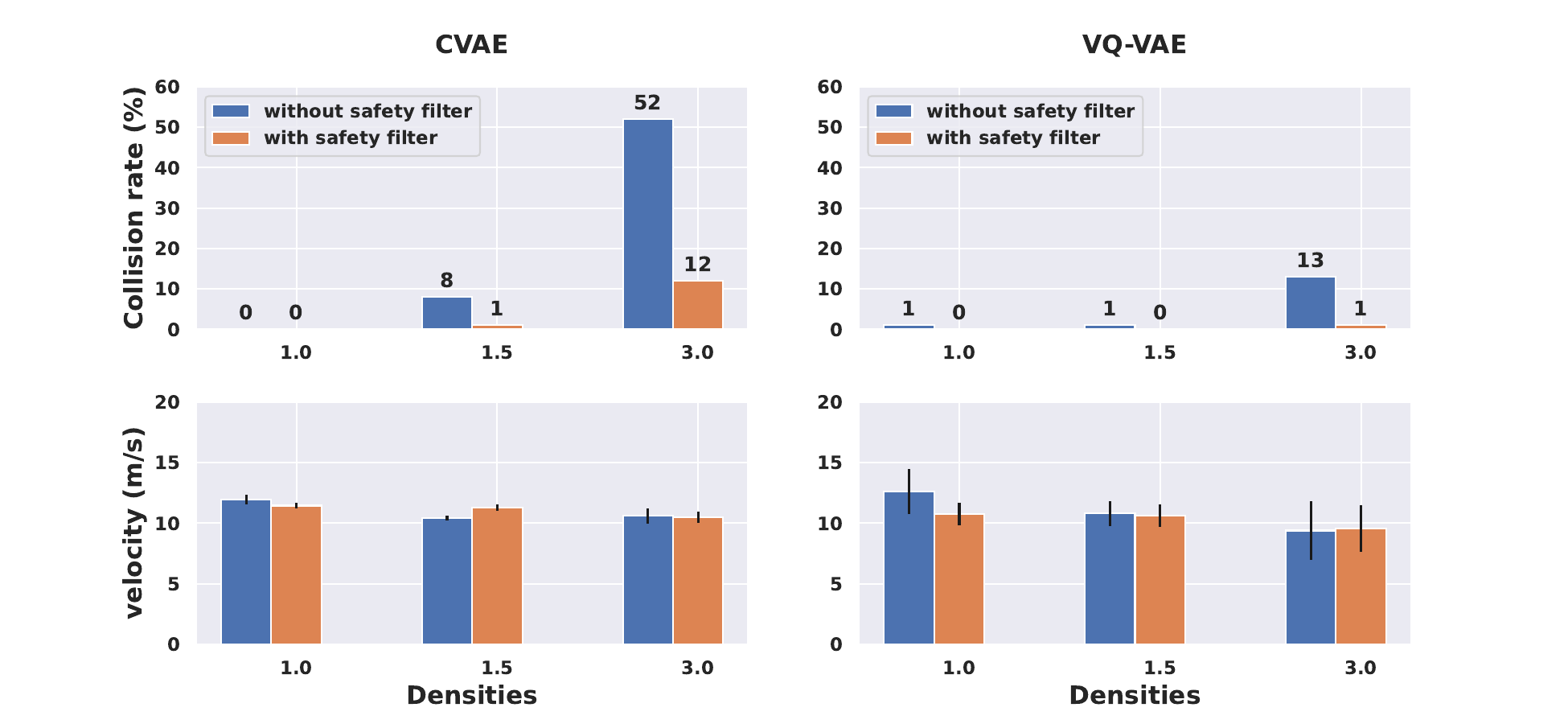}
\caption{Comparison of CVAE baseline \cite{shrestha2023end} and our VQ-VAE approach  with and without the safety filter. The superior samples generated from VQ-VAE allows for safe navigation even without the safety layer in some less/moderately dense scenarios}
\label{proj_layers}
\vspace{-0.2cm}
\end{figure}

\noindent Fig. \ref{proj_layers} illustrates the significance of a safety filter. We perform experiments for CVAE baseline \cite{shrestha2023end} and our VQ-VAE approach as both of them have similar safety with minor subtle differences that our proposed safety filter is based on barrier function whose parameters are also learned. The results are summarized in Fig.\ref{proj_layers}. It can be seen that up to traffic density 1.5, the multi-modal sampling ensured by VQ-VAE itself is enough to maintain a low collision-rate. In contrast, the CVAE-baseline without safety filter demonstrates an eight times higher collision-rate. At higher densities both CVAE and VQ-VAE based approaches require the explicit constraint handling provided by the safety filter to ensure low collision-rate. It can also be seen that the incorporation of safety filter typically makes the vehicle more conservative by reducing its forward speed. 


\section{Conclusion and future works}
We demonstrated how we can train a VQ-VAE model to learn a discrete prior over the space of optimal trajectories. Moreover, we can train a PixelCNN to sample from the learned discrete latent space. We showed that VQ-VAE is more capable of capturing the inherent multi-modality in driving behaviours and thus demonstrates superior performance than CVAE based approaches. In particular, we compare against the SOTA approach of \cite{shrestha2023end} and showed a reduction of up to 12 times in collision-rate. We also proposed a learnable safety filter that uses the concept of barrier function to enhance safety in dense traffic scenarios. The safety layer is trained in a self-supervised manner. Our future efforts are geared towards extending our approach to navigation in unstructured dynamic environments like human crowds and manipulation for high-dimensional systems.

\vspace{-0.1cm}


\begin{table*}[!t]
\centering
\caption{\scriptsize{List of Inequality Constraints Used in the Safety Filter}}
\small
\begin{tabular}{|c|c|c|c|c|c|}
\hline
Constraint Type & Expression & Parameters   \\ \hline
\makecell{Discrete-time Barrier \\ for Collision Avoidance at step $k$\\ $g_{1} = (g_{1, 1},  g_{1, 2}, \dots g_{1, m} )$}  &\makecell{$g_{1, i }[k] : h_{1, i}[k+1]-h_{1, i}[k]>-\gamma_{obs} h_{1,i}[k]$, \\   $h_{1, i}[k] = -\frac{(x[k]-x_{o, i}[k])^2}{a^2}-\frac{(y[k]-y_{o, i}[k])^2}{b^2}+1$} & \makecell{$\frac{a}{2}, \frac{b}{2}$: dimension of combined ellipitcal\\ footprint of vehicle and obstacle . \\ $x_{o,i}[k], y_{o, i}[k]$: trajectory way-point of \\ obstacles at time step $k$\\ $m$: number of obstacles } \\ \hline
\makecell{Velocity bounds at step $k$ \\ $g_{2} = (g_{2, lb}, g_{2, ub} )$} & \makecell{$ g_{2, ub }[k]: \sqrt{\dot{x}[k]^2+\dot{y}[k]^2}\leq v_{max}$\\$g_{2, lb}[k]:\sqrt{\dot{x}[k]^2+\dot{y}[k]^2}\geq v_{min}$} & \makecell{$v_{min}, v_{max}$: min/max velocity\\ of the ego-vehicle}   \\ \hline
\makecell{Acceleration bounds at step $k$ \\$g_3$} & $g_{3}[k]: \sqrt{\ddot{x}[k]^2+\ddot{y}[k]^2}\leq a_{max}$ & \makecell{$a_{max}$: max acceleration \\of the ego-vehicle}  \\ \hline
\makecell{Discrete-time barrier \\ for Lane boundary at step $k$ \\ $g_{4} = (g_{4, lb}, g_{4, ub} ) $} & \makecell{$g_{4, ub}[k]: h_{4, ub}[k+1]-h_{4, ub}[k]\geq -\gamma_{lane} h_{4, ub}[k]$, \\ $g_{4, lb}[k]: h_{4, lb}[k+1]-h_{4, lb}[k]\geq -\gamma_{lane} h_{4, ub}[k]$\\ $h_{4, ub}[k] = -y[k]+y_{ub}$ \\ $h_{4, lb}[k] = y[k]-y_{lb}$} & \makecell{$y_{lb}, y_{ub}$: Lane limits} \\ \hline
\end{tabular}
\normalsize
\label{ineq_list}
\vspace{-0.6cm}
\end{table*}

\section{Appendix}\label{appendix}
\noindent This section extends the formulation of \cite{shrestha2023end}. Specifically, we incorporate collision and lane boundary constraints in terms of barrier functions, while retaining the differentiability and efficiency of the resulting optimization steps.
\subsection{Reformulating Constraints:} 
\noindent Table \ref{ineq_list} presents the list of all the constraints included in our projection optimizer. The collision avoidance, velocity and acceleration constraints presented there can be re-written in the following polar form by introducing additional variables.

\small
\begin{align}
    \textbf{f}_{o, i} = \left \{ \begin{array}{lcr}
x[k] -x_{o, i}[k]-d_{o, i}[k]\cos\alpha_{o, i}[k] \\
y[k] -y_{o, i}[k]-d_{o, i}[k]\sin\alpha_{o, i}[k] \\ 
\end{array} \right \} d_{o, i}[k]\geq 1
\label{sphere_proposed}
\end{align}
\normalsize
\vspace{-0.1cm}

\vspace{-0.3cm}

\small
\begin{align}
    \textbf{f}_{v} = \left \{ \begin{array}{lcr}
\dot{x}[k] -d_{v}[k]\cos\alpha_{v}[k] \\
\dot{y}[k] -d_{v}[k]\sin\alpha_{v}[k]\\ 
\end{array} \right \}, v_{min}\leq d_{v}[k]\leq v_{max}
\label{vel_bound_proposed}
\end{align}
\normalsize

\vspace{-0.5cm}
\small
\begin{align}
    \textbf{f}_{a} = \left \{ \begin{array}{lcr}
\ddot{x}[k] -d_{a}[k]\cos\alpha_{a}[k] \\
\ddot{y}[k] -d_{a}[k]\sin\alpha_{a}[k]\\ 
\end{array} \right \}, 0\leq d_{a}[k]\leq a_{max}
\label{acc_bound_proposed}
\end{align}
\normalsize

The variables $\alpha_{o, i}[k]]$, $\alpha_{v}[k]$, $\alpha_{a}[k]$, $d_{o, i}[k]$, $d_{v, i}[k]$, and $d_{a, i}[k]$  will be computed by the safety layer.

\subsubsection*{Incorporating obstacle barrier constraints based on \cite{adajania2023amswarm}} In \eqref{sphere_proposed}, $d_{o,i}[k]=1$ represents the boundary of the feasible set of the collision avoidance constraints in Table \ref{ineq_list}. In other words $d_{o,i}[k]=1$ ensures $h_{1,i}[k]=0$. Similarly, $d_{o,i}[k]>1$ signify the interior of the set. With this insight, we can define the polar reformulation of barrier constraints in the following form:

\begin{align}
    d_{o, i}[k]\geq 1 + (1-\gamma_{obs})(d_{o, i}[k-1] -1), \forall k
    \label{barrier_d}
\end{align}


The constraints \eqref{sphere_proposed} and \eqref{barrier_d}  differ only in the feasible region definition of $d_{o, i}[k]$. When $\gamma_{obs} =1 $, the constraints are equivalent.

We integrate \eqref{barrier_d} into our algorithm through a minor modification in the feasible region definition of $d_{o, i}[k]$. Let $^{t}d_{o, i}[k]$ be the value of $d_{o, i}[k]$ obtained at the $\it {t}$-th iteration of our optimization algorithm (presented later). We use it to approximate the lower bound on $d_{o, i}[k]$ for obstacle barrier constraints at the $\it {(t+1)}$-th iteration as 
\begin{align}
    d_{o, i}[k]\geq 1 + (1-\gamma_{obs})(^{t}d_{o, i}[k-1] -1)
    \label{barrier_bound}
\end{align}

\noindent The right-hand side of \eqref{barrier_bound} is constant, and thus the feasible region of $d_{o, i}[k]$ for barrier constraints is approximated through a simple lower bound. 

\subsubsection{Incorporating lane barrier constraints}
The standard form of lane constraints is as follows:
\small
\begin{align}
    y_{lb} \leq y[k] \leq y_{ub} \forall k
\end{align}
\normalsize
where $y_{lb}, y_{ub}$ are the lane limits. The discrete-time control barrier functions corresponding to these are given in \ref{ineq_list}: 

Consequently, the lane upper bound barrier constraints are,
\small
\begin{align}
    y[k+1] + (\gamma_{lane}-1)y[k] \leq \gamma_{lane} y_{ub} \forall k
\end{align}
\normalsize

In terms of trajectory parametrization \eqref{param}, this becomes:
\begin{align}
    \mathbf{W}[k+1]\mathbf{c_{y}} + (\gamma_{lane} -1)\mathbf{W}[k]\mathbf{c_{y}} \leq \gamma_{lane} y_{ub} \forall k
\end{align}
\normalsize
where $\mathbf{W}[k]$ represents the $k^{th}$ row of the $\mathbf{W}$ matrix and $\mathbf{c_{y}}$ is the coefficient vector for $\mathbf{y}$.  Since we want a combined inequality corresponding to all the time steps $k=0,1,2,...,n-1$, we can rewrite the above as :

\small
\begin{align}
    { \begin{array}{lcr}
\mathbf{W}_{1, ub}\mathbf{c_{y}} + (\gamma_{lane} -1)\mathbf{W}_{0, ub}\mathbf{c_{y}} \leq \mathbf{b}_{ub} \ or,\\ \\
(\mathbf{W}_{1, ub} + (\gamma_{lane} -1)\mathbf{W}_{0, ub})\mathbf{c_{y}} \leq \mathbf{b}_{ub}
\end{array}} 
\end{align}
\normalsize
where $\mathbf{W}_{1, ub} := \mathbf{W}[1:n-1]$, $\mathbf{W}_{0, ub}:= \mathbf{W}[0:n-2]$ and $\mathbf{b}_{ub}:= \gamma_{lane} y_{ub} \mathbf{1}_{(n-1) \times 1}$. \\
Define $\mathbf{G}_{ub} := \mathbf{W}_{1, ub} + (\gamma_{lane} -1)\mathbf{W}_{0, ub}$. This results in the following inequality constraint:

\small
\begin{align}
    \mathbf{G}_{ub}\mathbf{c_{y}} \leq \mathbf{b}_{ub} \label{upper-bound}
\end{align}
\normalsize

Similarly, we get the lane lower bound barrier constraints as:
\small
\begin{align}
    \mathbf{G}_{lb}\mathbf{c_{y}} \leq \mathbf{b}_{lb} \label{lower-bound}
\end{align}
\normalsize
where $\mathbf{G}_{lb} := \mathbf{W}_{1, lb} + (1 - \gamma_{lane})\mathbf{W}_{0, lb}$, $\mathbf{W}_{1, lb} := -\mathbf{W}[1:n-1]$, $\mathbf{W}_{0, lb}:= \mathbf{W}[0:n-2]$ and $\mathbf{b}_{lb}:= -\gamma_{lane} y_{lb} \mathbf{1}_{(n-1) \times 1}$.

\subsubsection{Reformulated Problem} Using the developments in the previous section and the trajectory parametrization presented in \eqref{param}, we can now replace the safety layer optimization \eqref{projection_cost}-\eqref{projection_const} with the following. Note the subscript $j$ that signifies the problem is defined for projecting the $j^{th}$ sampled trajectory.

\vspace{-0.6cm}
\small
\begin{subequations}
\begin{align}
    \overline{\boldsymbol{\xi}}_j^{*} = \arg\min_{\overline{\boldsymbol{\xi}}^*_j}\frac{1}{2}\Vert \overline{\boldsymbol{\xi}}^*_j-{\boldsymbol{\xi}}_j^*\Vert_2^2\label{cost_reform}  \\
    \textbf{A} \overline{\boldsymbol{\xi}}^*_j= \textbf{b} \label{eq_reform} \\
    \widetilde{\textbf{F}} \hspace{0.05cm} \overline{\boldsymbol{\xi}}^*_j = \widetilde{\textbf{e}}(\boldsymbol{\alpha}_j, \textbf{d}_j) \label{nonconvex_reform}  \\
    \textbf{d}_{min} \leq \textbf{d}_j\leq \textbf{d}_{max} (\gamma_{obs}) \label{d_reform_1}\\
     \textbf{G}_{\gamma_{lane}}\overline{\boldsymbol{\xi}}^*_j \leq \textbf{y}_{lane} \label{lane_reform}
\end{align}
\end{subequations}
\normalsize
\vspace{-0.5cm}
\small
\begin{align}
    \widetilde{\textbf{F}} = \begin{bmatrix}
    \begin{bmatrix}
    \textbf{F}_{o}\\
    \dot{\textbf{W}}\\
    \ddot{\textbf{W}}
    \end{bmatrix} & \textbf{0}\\
    \textbf{0} & \begin{bmatrix}
    \textbf{F}_{o}\\
    \dot{\textbf{W}}\\
    \ddot{\textbf{W}}
    \end{bmatrix} 
    \end{bmatrix}, \widetilde{\textbf{e}} = \begin{bmatrix}
    \textbf{x}_o+a \textbf{d}_{o, j}\cos\boldsymbol{\alpha}_{o, j}\\
     \textbf{d}_{v, j}\cos\boldsymbol{\alpha}_{v, j}\\
  \textbf{d}_{a, j}\cos\boldsymbol{\alpha}_{a, j}\\
 \textbf{y}_o+a \textbf{d}_{o, j}\sin\boldsymbol{\alpha}_{o, j}\\
     \textbf{d}_{v, j}\sin\boldsymbol{\alpha}_{v, j}\\
  \textbf{d}_{a, j}\sin\boldsymbol{\alpha}_{a, j}\\
    \end{bmatrix},
\end{align}
\vspace{-0.3cm}
\begin{align}
    \textbf{G}_{\gamma_{lane}} = \begin{bmatrix}
        \textbf{G}_{ub} & \textbf{G}_{lb}
    \end{bmatrix}^T, \textbf{y}_{lane} = \begin{bmatrix}
        \textbf{b}_{ub} & \textbf{b}_{lb}
    \end{bmatrix}^T
\end{align}
\vspace{-0.3cm}
\begin{align*}
    \boldsymbol{\alpha}_j = (\boldsymbol{\alpha}_{o, j}, \boldsymbol{\alpha}_{a,j}, \boldsymbol{\alpha}_{v,j}), \qquad \textbf{d}_j =  (\textbf{d}_{o, j}, \textbf{d}_{v, j}, \textbf{d}_{a, j})
\end{align*}
\normalsize

\noindent Constraints \eqref{nonconvex_reform}-\eqref{lane_reform} acts as substitutes for $\textbf{g}(\overline{\boldsymbol{\xi}}_j^*)\leq 0 $ in the optimization \eqref{projection_cost}-\ref{projection_const}. 

We form matrix $\textbf{F}_o$ by stacking the matrix $\textbf{W}$ from (\ref{param}) as many times as the number of neighbouring vehicles considered for collision avoidance at a given planning cycle. The vector $\textbf{x}_o, \textbf{y}_o$ is formed by appropriately stacking $x_{o, i}[k], y_{o, i}[k]$ at different time instants and for all the neighbours. Similar construction is followed to obtain $\boldsymbol{\alpha}_{o, j}, \boldsymbol{\alpha}_{v, j}, \boldsymbol{\alpha}_{a, j}, \textbf{d}_{o, j}, \boldsymbol{d}_{v, j} \boldsymbol{d}_{a, j}$. The vector $\textbf{y}_{lane}$ is formed by stacking the upper and lower lane bounds after repeating them $n$ times (planning horizon). Similarly,  vectors $\mathbf{d}_{min}, \mathbf{d}_{max}$ are formed by stacking the lower and upper bounds for $\textbf{d}_{o, j}, \boldsymbol{d}_{v, j} \boldsymbol{d}_{a, j}$ (recall \eqref{sphere_proposed}-\eqref{acc_bound_proposed}). Note that the upper bound for $\textbf{d}_{o, j}$ can be simply some large number, while the lower bound depends on $\gamma_{obs}$ (recall \eqref{sphere_proposed}, \eqref{barrier_bound}). Moreover, these bounds are the same across all batches.

\begin{remark}
    The vector $\boldsymbol{\gamma} = (\gamma_{lane}, \gamma_{obs})$ form the learnable barrier function parameters introduced in \eqref{projection_const}, Section \ref{LSF}
\end{remark}

\subsubsection{Solution Process} We relax the non-convex equality \eqref{nonconvex_reform} and affine inequality constraints as $l_2$ penalties and augment them into the projection cost \eqref{cost_reform}.

\small
\begin{dmath}
    \mathcal{L} = \frac{1}{2}\left\Vert \overline{\boldsymbol{\xi}}^*_j-\boldsymbol{\xi}^*_j\right\Vert_2^2- \boldsymbol{\lambda}_{j}^T \overline{\boldsymbol{\xi}}^*_j+\frac{\rho}{2} \left \Vert \widetilde{\textbf{F}} \overline{\boldsymbol{\xi}}^*_j-\widetilde{\textbf{e}}\right \Vert_2^2+  \frac{\rho}{2}\left \Vert \mathbf{G}_{\gamma_{lane}} \overline{\boldsymbol{\xi}}^*_{j} - \textbf{y}_{lane} + \mathbf{s}_j \right \Vert^2 = \frac{1}{2}\left\Vert \overline{\boldsymbol{\xi}}^*_j-\boldsymbol{\xi}^*_j\right\Vert_2^2-\boldsymbol{\lambda}_{j}^T \overline{\boldsymbol{\xi}}^*_j+\frac{\rho}{2} \left \Vert \textbf{F} \overline{\boldsymbol{\xi}}^*_j-\textbf{e}\right \Vert_2^2
    \label{aug_lag}
\end{dmath}
\normalsize
\begin{align}
    \textbf{F} = \begin{bmatrix}
        \widetilde{\textbf{F}}\\
        \textbf{G}_{\gamma_{lane}}
    \end{bmatrix}, \textbf{e} = \begin{bmatrix}
        \widetilde{\textbf{e}}\\
        \textbf{y}_{lane}-\textbf{s}_j
    \end{bmatrix}
\end{align}

\noindent Note, the introduction of the Lagrange multiplier $\boldsymbol{\lambda}$ that drives the residual of the second and third quadratic penalties to zero \cite{split_bergman}. We minimize \eqref{aug_lag} subject to \eqref{eq_reform} through Alternating Minimization (AM), which reduces to the following steps \cite{masnavi2022visibility}, wherein the left superscript $t$ represents the iteration index.

\vspace{-0.6cm}
\small
\begin{subequations}
    \begin{align}
        {^{t+1}\boldsymbol{\alpha}_j} = \arg\min_{\boldsymbol{\alpha}_j} \mathcal{L}({^t}\overline{\boldsymbol{\xi}}_j^*, {^t}\textbf{d}_j, \boldsymbol{\alpha}_j {^t}\boldsymbol{\lambda}_j, {^t}\textbf{s}_j ) \label{am_alpha}\\
        {^{t+1}\textbf{d}_j} = \arg\min_{\textbf{d}_j} \mathcal{L}({^t}\overline{\boldsymbol{\xi}}_j^*, \textbf{d}_j, {^{k+1}}\boldsymbol{\alpha}_j, {^t}\boldsymbol{\lambda}_j, {^t}\textbf{s}_j) \label{am_d} \\ 
        {^{t+1}}\mathbf{s} =\text{max}\left(0, -\mathbf{G}_{\gamma_{lane}} {^{t}}\overline{\boldsymbol{\xi}}_{j}^* - \textbf{y}_{lane}\right) \label{am_s} \\
        {^{t+1}}\boldsymbol{\lambda}_j = {^{t}}\boldsymbol{\lambda}_j+\rho\textbf{F}^T (\textbf{F}\hspace{0.05cm} {^t}\overline{\boldsymbol{\xi}}_j^*-{^{t}}\textbf{e}_j  )\label{am_lambda} \\
        {^{t+1}}\textbf{e}_j = \begin{bmatrix}
        \widetilde{\textbf{e}} ({^{t+1}} \boldsymbol{\alpha}_j, {^{t+1}}\textbf{d}_j ) \label{am_e} \\
        \textbf{y}_{lane}-{^{t+1}}\textbf{s}_j
    \end{bmatrix}\\
        {^{t+1}}\overline{\boldsymbol{\xi}}_j^* = \arg\min_{\overline{\boldsymbol{\xi}}_j^*}\mathcal{L}(\overline{\boldsymbol{\xi}}_j^*, {^{t+1}}\boldsymbol{\lambda}_j, {^{t+1}}\textbf{e}_j ) \label{am_xi}
    \end{align}
\end{subequations}
\normalsize

As can be seen, we optimize over only one group of variables at each AM step  while others are held fixed at values obtained at the previous updates. Steps \eqref{am_lambda}-\eqref{am_e} have a closed-form solution, while \eqref{am_xi} is an equality constrained QP \cite{shrestha2023end}. Thus, we can trace the gradients through the unrolled optimization step and train the safety filter proposed in Section \ref{LSF}. Moreover, as shown in \cite{shrestha2023end}, the QP \eqref{am_xi} is easily batchable and can be made free of matrix factorization.

\bibliography{ref_iros_2023}
\bibliographystyle{IEEEtran}

\end{document}